\documentclass{article}

\usepackage{amsmath}
\usepackage{algorithm}
\usepackage{algorithmic}
\usepackage{graphicx}
\usepackage{multirow}
\usepackage{graphicx}
\usepackage{subcaption}
\usepackage{url}
\usepackage{todonotes}
\date{}

\title{Scalable Time-Series Causal Discovery with Approximate Causal Ordering}

\author{Ziyang Jiao, Ce Guo and Wayne Luk\\
Department of Computing\\
Imperial College London\\
\{ziyang.jiao23, c.guo, w.luk\}@imperial.ac.uk}

\begin{document}
\maketitle

\abstract{Causal discovery in time-series data presents a significant computational challenge. Standard algorithms are often prohibitively expensive for datasets with many variables or samples. This study introduces and validates a heuristic approximation of the VarLiNGAM algorithm to address this scalability problem. The standard VarLiNGAM method relies on an iterative search, recalculating statistical dependencies after each step. Our heuristic modifies this procedure by omitting the iterative refinement. This change permits a one-time precomputation of all necessary statistical values. The algorithmic modification reduces the time complexity from $O(m^3n)$ to $O(m^2n + m^3)$ while keeping the space complexity at $O(m^2)$, where $m$ is the number of variables and $n$ is the number of samples. While an approximation, our approach retains VarLiNGAM's essential structure and empirical reliability. On large-scale financial data with up to 400 variables, our algorithm achieves a 7--13x speedup over the standard implementation and a 4.5x speedup over a GPU-accelerated version. Evaluations across medical imaging, web server monitoring, and finance demonstrate the heuristic's robustness and practical scalability. This work offers a validated balance between computational efficiency and discovery quality, making large-scale causal analysis feasible on personal computers.
}

\newcommand{\ceguo}[1]{\todo[color=cyan!30,inline]{Ce: #1}}

\section{Introduction}

Time-series causal discovery is the process of inferring cause-and-effect relationships from data points recorded in chronological order. The goal is to determine how variables influence one another, both at the same time (contemporaneous effects) and across different times (lagged effects). While critical in fields from finance to climate science, the application of these methods to the large datasets common in modern industry is often prohibitively slow.

An example is the well-regarded VarLiNGAM algorithm, whose iterative nature results in a computational complexity of $O(m^{3}n)$, creating a severe scalability bottleneck for datasets with many variables ($m$) or samples ($n$). To address this challenge, this study introduces and validates a novel heuristic approach based on VarLiNGAM.

Our method intentionally modifies the standard iterative procedure by replacing the costly, step-by-step causal ordering refinement with a highly efficient, one-time precomputation of all necessary statistical values. This algorithmic change is based on the central hypothesis that for many complex time-series, the initial dependency structure contains sufficient information to identify the correct causal ordering without iterative updates. This change reduces the complexity to $O(m^{2}n+m^{3})$. While this approach is an approximation and thus sacrifices a degree of theoretical exactness, it retains the essential structure of the original algorithm and, as our experiments show, its empirical reliability.

Our method involves an efficient precomputation method. Precomputation has been shown to be a useful technique for performance enhancement in data analysis. For example,  it has been reported that preprocessing for approximate Bayesian computation in image analysis can reduce the average runtime required for model fitting from 71 hours to 7 minutes \cite{moores2015pre}.

The key contributions of this work are as follows:
\begin{enumerate}
    \item A computational bottleneck analysis of the VarLiNGAM algorithm, identifying the iterative data refinement within its DirectLiNGAM estimator as the primary source of its $O(m^3n)$ complexity.

    \item The design and analysis of a novel heuristic, which approximates the standard procedure by replacing iterative refinement with an approximation and precomputation strategy. This reduces the theoretical time complexity to $O(m^2n + m^3)$.

    \item An evaluation of the proposed heuristic on diverse synthetic and real-world datasets. The results demonstrate significant speedups (up to 13x over the official CPU implementation and 4.5x over a GPU version) with a negligible cost to discovery accuracy.
\end{enumerate}

On large-scale financial data with up to 400 variables, our algorithm achieves a 7 to 13 times speedup over the official implementation \cite{ikeuchi2023python} and an approximate 4.5 times speedup over a GPU-accelerated version \cite{akinwande2024acceleratedlingam}. This work offers a validated balance between computational efficiency and causal discovery quality, extending the feasibility of applying causal causal discovery to large-scale, real-world problems using standard hardware resources. The source code of the proposed approach is available online\footnote{Code repository: \url{https://github.com/ceguo/varlingam-heuristic}}.

\section{Background and Related Work}

Causal discovery from time-series data is concerned with inferring directed causal graphs from multivariate observational data. This task is distinct from analysis in the independent and identically distributed (i.i.d.) setting because it must explicitly account for temporal dependencies, where a variable at one point in time can influence another variable at a future point in time \cite{assaad2022survey}. The key challenges in this domain include correctly identifying the duration of time lags, managing potential feedback loops and cyclical relationships, and separating direct causal influences from indirect correlations mediated by other variables. The typical output of these methods is a directed graph that provides a map of a system's causal mechanisms. Such graphs are valuable in many fields, from financial computing and policy analysis to understanding functional connectivity in the brain \cite{chu2008search, seth2015granger}.

\subsection{Types of Time-Series Causal Discovery Methods}


A foundational approach is Granger causality, first proposed by Granger in 1969 \cite{granger2001investigating}. The core idea is that a time series \(X\) is said to be a Granger-cause of another time series \(Y\) if the past values of \(X\) contain information that helps predict the future values of \(Y\) better than using only the past values of \(Y\). While originally formulated for bivariate, linear systems, this concept has been adapted to handle more complex scenarios. Variants include multivariate Granger causality \cite{arize1993determinants}, which considers the influence of multiple variables simultaneously, and conditional Granger models \cite{chen2004analyzing}, which can account for the confounding influence \cite{geweke1982measurement} of other time series.

Information-theoretic methods provide a non-parametric alternative that can capture nonlinear relationships. These approaches are based on concepts from information theory, such as entropy, which measures the uncertainty of a variable. A key metric is Transfer Entropy (TE), which quantifies the reduction in uncertainty about a variable's future state given the past state of another variable \cite{schreiber2000measuring, bossomaier2016transfer}. Other related metrics include mutual information (MI) and conditional mutual information (CMI), which measure the statistical dependence between variables \cite{kraskov2004estimating}. For Gaussian variables, it has been shown that Granger causality and Transfer Entropy are mathematically equivalent \cite{barnett2009granger}. A common challenge for these methods, however, is that the symmetric nature of the measures can make it difficult to determine the direction of the causal link without additional assumptions.

Constraint-based methods infer causal structure by conducting a series of conditional independence (CI) tests. These methods, which originate from the i.i.d. setting with algorithms like the Peter-Clark (PC) algorithm \cite{spirtes2001causation} and Fast Causal Inference (FCI) \cite{kalisch2007estimating}, are adapted for time series by including lagged variables in the conditioning sets of the CI tests. Algorithms such as PCMCI \cite{runge2019detecting} and tsFCI \cite{entner2010causal} have been developed for this purpose. They typically start with a fully connected graph and iteratively remove edges between variables that are found to be conditionally independent, eventually revealing the underlying causal skeleton.

This paper focuses on function-based methods. These methods assume a specific data-generating process. By imposing structural constraints on the relationships between variables, these models can often identify a unique causal graph where other methods might only identify a class of equivalent graphs.

A prominent family of function-based methods is the Linear Non-Gaussian Acyclic Model (LiNGAM). LiNGAM-based methods assume that the causal relationships between variables are linear, the system is acyclic, and the external noise sources affecting each variable are independent and non-Gaussian \cite{shimizu2006linear, shimizu2014lingam}. The non-Gaussianity assumption is critical, as it breaks the statistical symmetry that makes linear Gaussian models non-identifiable. Under these assumptions, the causal structure can be identified using techniques like Independent Component Analysis (ICA) \cite{hyvarinen2004independent}. The Vector Autoregressive LiNGAM (VarLiNGAM) model extends this framework to time-series data \cite{hyvarinen2010estimation}. It works by first fitting a standard Vector Autoregressive (VAR) model to account for the time-lagged causal influences. It then applies the LiNGAM algorithm to the residuals of the VAR model to discover the contemporaneous, or instantaneous, causal structure.

Other function-based models have different assumptions on the function's form. For example, Additive Noise Models (ANM) \cite{hoyer2008nonlinear, mooij2009regression, nowzohour2015score} can handle nonlinear causal relationships, provided that the noise is additive and independent of the causes. Post-Non-Linear (PNL) \cite{zhang2012identifiability} causal models further generalize this by allowing an additional nonlinear transformation of the effect variable. In the time-series context, methods like DYNOTEARS \cite{pamfil2020dynotears} have been proposed for learning dynamic structures under a continuous optimization framework.

The VarLiNGAM framework is chosen for this work due to its ability to identify a full causal structure under a well-understood set of assumptions that are met in many real-world domains.

\subsection{Scalability and Acceleration}

A common problem across all families of causal discovery methods is their computational efficiency, which often limits their application to datasets with a large number of variables or time points. In response, a significant body of research focus on acceleration, which can be broadly categorized into hardware-centric and algorithmic approaches.

Hardware-centric acceleration aims to reduce the execution time of existing algorithms by using specialized processors. GPU acceleration is a common strategy. For constraint-based methods, GPUs have been used to parallelize the large number of required conditional independence tests \cite{zarebavani2020cupc, hagedorn2021gpu}. For function-based methods, GPUs have been used to accelerate the intensive matrix operations involved in algorithms like LiNGAM \cite{akinwande2024acceleratedlingam, shahbazinia2023paralingam}. For even larger-scale problems, some work has explored the use of supercomputers to distribute the workload across thousands of nodes \cite{matsuda2022accelerating}. Other research has focused on using Field-Programmable Gate Arrays (FPGAs) to create custom hardware pipelines for specific bottlenecks, such as the generation of candidate condition sets for CI tests \cite{guo2022accelerating, guo2023fpga, guo2023codesign}. These hardware-centric solutions are effective but depend on the availability of specialized and often costly computing resources.

In contrast, our work explores a purely algorithmic path to scalability. Instead of using more computational resources to execute the same number of operations faster, we modify the algorithm itself to fundamentally reduce the total operation count. This makes our contribution distinct from, and complementary to, existing work on hardware acceleration. Our focus is on improving performance on standard, widely accessible hardware, which is a different but equally important direction for making large-scale causal discovery more practical for a broader community of researchers and practitioners.

\section{Bottleneck Analysis of DirectLiNGAM}
\label{sec:bottleneck}

As part of the VarLiNGAM procedure, DirectLiNGAM is the de-facto method to find the causal ordering and contemporaneous causal graph \cite{shimizu2011directlingam}. It operates iteratively, identifying and removing the most exogenous variable from a set of candidates in each pass. This iterative refinement is both the source of its accuracy and its high computational cost.

The DirectLiNGAM algorithm is designed to find the causal ordering of variables through an iterative search procedure. The time complexity of VarLiNGAM is equivalent to that of DirectLiNGAM, due to the high efficiency of VAR. Each main loop of DirectLiNGAM identifies the most exogenous variable among a set of current candidates. The following is a detailed breakdown of the steps performed within a single loop to find the \(k\)-th variable in the causal ordering, \(c_k\), along with an analysis of the computational cost of each step. In this analysis, \(m_k\) denotes the number of remaining variables at the start of the iteration, and \(n\) is the number of samples.

\begin{enumerate}
    \item Standardization: First, the current data matrix \(X^{(k-1)}\), which contains the \(m_k\) variables yet to be ordered, is standardized so that each column has a mean of zero and a variance of one. This ensures that the scale of the variables does not affect the subsequent calculations.

    Execution Time: This step requires calculating the mean and standard deviation for each of the \(m_k\) columns. Both operations have a complexity of \(O(n)\) for a single column. Therefore, the total time complexity for standardizing the entire matrix is \(O(m_k \cdot n)\).

    \item Pairwise Residual Calculation: For every pair of variables (\(x_i, x_j\)) with indices in the current set \(U^{(k-1)}\), the linear regression residual is computed. The residual \(r_{i \leftarrow j}\) represents the part of \(x_i\) that cannot be linearly explained by \(x_j\). It is calculated as:
    \begin{equation}
     r_{i \leftarrow j} = x_i - \frac{\text{cov}(x_i, x_j)}{\text{var}(x_j)} x_j
    \end{equation}

    Execution Time: This is a computationally intensive step. For each of the \(O(m_k^2)\) pairs of variables, calculating the covariance and variance takes \(O(n)\) time, and the subsequent vector operations also take \(O(n)\) time. Consequently, the total time complexity for this step is \(O(m_k^2 \cdot n)\).

    \item Scoring via Mutual Information: A measure of dependence, \(T_{i \leftarrow j}\), is calculated for each pair of variables. This score approximates the mutual information between a variable and its residual after regressing on another. Using an entropy approximation \(H(\cdot)\), it is defined as:
    \begin{equation}
     T_{i \leftarrow j} = H(x_i) - H(r_{i \leftarrow j})
    \end{equation}
    A lower value of \(T_{i \leftarrow j}\) indicates that \(x_j\) explains less of the information in \(x_i\), suggesting \(x_i\) is more independent of \(x_j\).

    Execution Time: The entropy calculation for a single vector of length \(n\) has a complexity of \(O(n)\). This step requires calculating the entropy for all \(m_k\) variables and all \(O(m_k^2)\) residuals computed in the previous step. The total time complexity is therefore dominated by the calculation of residual entropies, resulting in a cost of \(O(m_k^2 \cdot n)\).

    \item Variable Selection: For each candidate variable \(x_i\), an aggregate score \(M_i\) is computed by summing a function of the pairwise scores against all other remaining variables \(x_j\):
    \begin{equation}
     M_i = \sum_{j \in U^{(k-1)}, j \neq i} f(T_{i \leftarrow j}, T_{j \leftarrow i})
     \label{equ:mi}
    \end{equation}
    The variable \(c_k\) with the score indicating maximum overall independence is selected as the \(k\)-th variable in the causal ordering.

    Execution Time: For each of the \(m_k\) variables, computing the aggregate score involves summing \(m_k-1\) terms. Assuming the function \(f\) is \(O(1)\), this takes \(O(m_k)\) time per variable. The total time to calculate all aggregate scores is \(O(m_k^2)\). This is computationally less significant compared to the previous steps.

    \item Iterative Data Refinement: This is the crucial step that ensures the correctness of subsequent iterations. The algorithm prepares the data matrix for the next loop, \(X^{(k)}\), by removing the influence of the just-found variable \(c_k\) from all other remaining variables. For each remaining variable index \(j\), the corresponding column in the new data matrix is updated with its residual:
    \begin{equation}
     x_j^{(k)} = r_{j \leftarrow c_k}^{(k-1)} = x_j^{(k-1)} - \frac{\text{cov}(x_j^{(k-1)}, x_{c_k}^{(k-1)})}{\text{var}(x_{c_k}^{(k-1)})} x_{c_k}^{(k-1)}
    \end{equation}
    The set of candidate indices is also updated, \(U^{(k)} = U^{(k-1)} \setminus \{c_k\}\), and the process repeats to find the next variable, \(c_{k+1}\).

    Execution Time: This step involves \(m_k-1\) residual calculations. Since each residual calculation takes \(O(n)\) time, the total time complexity for this refinement step is \(O(m_k \cdot n)\).
\end{enumerate}

The fifth step, iterative data refinement, is the fundamental bottleneck. Because the entire data matrix \(X\) is updated in every one of the \(m\) main iterations. In each iteration, all pairwise residuals and entropy calculations must be re-computed from scratch. This nested computational structure is what leads to the high overall complexity of \( O(m^3 n) \).

\section{Proposed Approach: Approximate Causal Ordering}\label{sec:methodology}

Our work focuses on accelerating the core bottleneck of the VarLiNGAM algorithm: the estimation of the instantaneous causal matrix \( B_0 \) using its default estimator, DirectLiNGAM \cite{shimizu2011directlingam}. To address this, we propose a novel heuristic approximation that modifies the causal ordering algorithm in DirectLiNGAM for VarLiNGAM.

\subsection{Motivation}
The VarLiNGAM algorithm employs a VAR model as its initial step to refine the data. This approach involves analyzing the dynamics of each variable in relation to its past values and those of other variables, thereby identifying the underlying influences that shape the time series.

By subtracting these past influences from the original data, the VAR model generates residuals that capture the unexpected or unforeseen events that occur at each point in time. These residuals serve as a proxy for the causal relationships between the variables, allowing the algorithm to focus on the simultaneous shocks rather than the complex time-series data.

The use of the VAR model as a preprocessing step has a significant impact on the subsequent analysis. By removing these past influences, the algorithm is effectively reduced to a problem of identifying relationships between the instantaneous shocks, rather than navigating the intricate web of causal links inherent in the raw data. Consequently, we propose that calculating relationships from clean VAR residuals offers a viable shortcut, using the VarLiNGAM framework's initial cleaning step to diminish the need for subsequent refinement.

Notably, this heuristic is specifically tailored to the context of time-series data, where the preprocessing step performed by the VAR model provides the necessary foundation for the algorithm. However, it remains unclear whether this approach would be effective in non-time-series scenarios, where such an initial cleaning step may not be available.

\subsection{Algorithmic Modification and Precomputation}

The implementation of our heuristic fundamentally alters the program flow. Instead of an iterative refinement process, it adopts a precompute-and-lookup strategy.

\begin{enumerate}
    \item Precomputation of Variable Entropies: Before the search for causal ordering begins, the entropy of each standardized column \(x_i\) from the original data matrix \(X\) is calculated once and stored in an array of size \(m\).
    
    \item Precomputation of Residual Entropies: All \(m \times (m-1)\) pairwise residuals, \(r_{i \leftarrow j}\) for all \(i \neq j\), are calculated from the single, original, unaltered data matrix \(X\). The entropy of each of these residuals is then computed and stored in an \(m \times m\) matrix.
    
    \item Accelerated Causal Ordering Search: The main loop to find the causal ordering proceeds for \(m\) iterations. However, in each iteration, it performs its search over the same, static set of precomputed entropy values. The scoring calculation (Step 3 and 4 of the original method) is reduced from a series of vector operations to a few memory lookups from the precomputed arrays. Crucially, the data matrix \(X\) is never updated.
\end{enumerate}

\begin{algorithm}
\caption{Proposed Heuristic Causal Order Search}
\begin{algorithmic}[1]
\STATE Input: Data matrix \(X\), initial set of indices \(U = \{1, ..., m\}\)
\STATE Output: Causal order \(K\)
\STATE \(E_x \leftarrow \text{precomputeVariableEntropies}(X)\)
\STATE \(E_r \leftarrow \text{precomputeResidualEntropies}(X)\)
\STATE \(K \leftarrow []\)
\FOR{\(k = 1\) to \(m\)}
    \STATE Find \(c_k \in U\) that minimizes the dependence score \(M\) in Equation~\ref{equ:mi} by looking up values in \(E_x\) and \(E_r\).
    \STATE Append \(c_k\) to \(K\).
    \STATE \(U \leftarrow U \setminus \{c_k\}\)
\ENDFOR
\STATE return \(K\)
\end{algorithmic}
\label{alg:fast}
\end{algorithm}

Algorithm~\ref{alg:fast} outlines our heuristic, where the expensive calculations are moved outside the main loop into a precomputation phase. The main loop  no longer contains any residual or entropy calculations, and most importantly, it lacks the data update step.

This algorithmic change directly impacts the complexity. The two precomputation steps have a combined complexity of \( O(m n + m^2 n) = O(m^2 n) \). The main search loop, which runs \(m\) times, now only performs \(O(m^2)\) work per iteration (for pairwise score lookups and comparisons), resulting in a total search complexity of \(O(m^3)\). The final complexity of our heuristic is the sum of these parts, \( O(m^2 n + m^3) \), which is substantially lower than the original's \( O(m^3 n) \), since the number of variables $m$ is typically of the order $10^{3}$ and beyond \cite{jeon2020using}. Also, since the approximation only needs to store the dependence score $M$ for each pair of variables, the space complexity is $O(m^2)$, which is the same as the original VarLiNGAM algorithm.

\section{Evaluation}\label{sec:evaluation}

\subsection{Experimental Setup}
To assess the performance of our heuristic, we use two standard metrics: Structural Hamming Distance (SHD) \cite{peters2014structural} and F1-score \cite{shimizu2011directlingam}. We conduct experiments on a variety of datasets to test performance under different conditions. To ensure a fair comparison of computational performance, both the original and our proposed implementations are developed in Python using identical numerical libraries such as NumPy and SciPy. No explicit multi-threading or other parallel frameworks are used in the CPU implementations, meaning that the observed speedup is attributable solely to the change in the algorithm's design. Since real-world causal discovery tasks are often executed on personal computers \cite{le2016fast}, we use a laptop for the evaluation. The laptop has an Intel Core Ultra 7 155H CPU and 32GB DDR5 memory without a dedicated GPU. For reference implementations that must run on GPUs, we use a different machine that contains an NVIDIA Tesla T4 GPU.

\subsection{Experiments with Synthetic Data}

\begin{figure}[t]
    \centering
    \subfloat[Execution Time (secs)]{
    \includegraphics[width=0.31\linewidth]{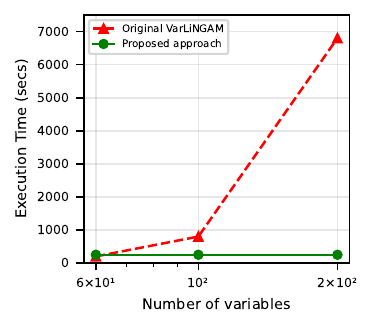}
    }
    \subfloat[Accuracy: Original]{
    \includegraphics[width=0.31\textwidth]{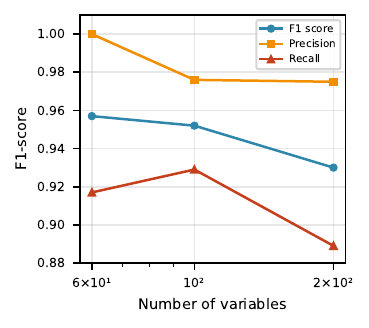}
    }
    \subfloat[Accuracy: Proposed]{
    \includegraphics[width=0.31\textwidth]{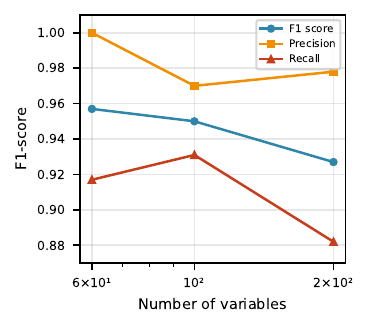}
    }
    \caption{Performance on synthetic data with a fixed sample size $n=10,000$ and varying number of variables.}
    \label{fig:simulated_results_fixn}
\end{figure}

\begin{figure}[t]
    \centering
    \subfloat[Execution Time (secs)]{
        \includegraphics[width=0.31\linewidth]{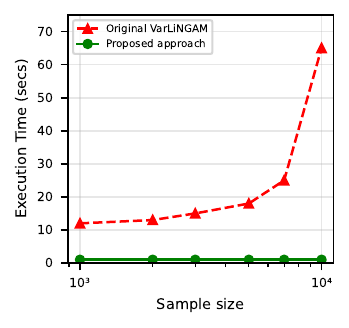}
    }
    \subfloat[Accuracy: Original]{
    \includegraphics[width=0.31\textwidth]{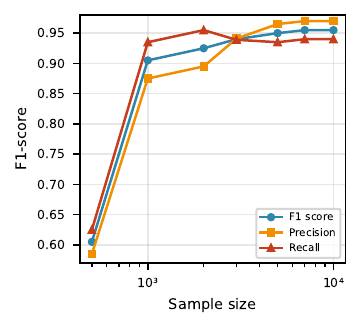}
    }
    \subfloat[Accuracy: Proposed]{
        \includegraphics[width=0.31\textwidth]{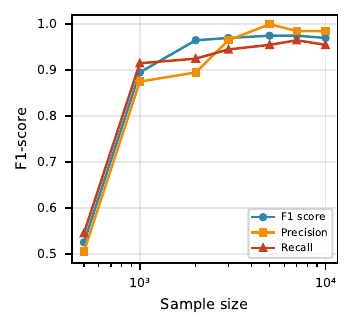}
    }

    \caption{Performance on synthetic data with a fixed number of variables $m=50$ and a varying number of samples.}
    \label{fig:simulated_results_fixm}
\end{figure}

We performed two sets of experiments on synthetic data to analyze the performance of our heuristic against the original algorithm under controlled conditions. The results, shown in Figure~\ref{fig:simulated_results_fixn} and Figure~\ref{fig:simulated_results_fixm}, illustrate the practical trade-offs between computation time and discovery accuracy.
\begin{itemize}
    \item Fixed sample size ($n=10,000$) with increasing number of variables. As shown in Figure \ref{fig:simulated_results_fixn}, the execution time of the original search step grows rapidly, which is consistent with its high computational complexity. In contrast, our heuristic's search time remains nearly constant. While the precomputation step introduces some overhead, the overall time saving is substantial. Note that a practical dataset may have a smaller sample size, so the acceleration can be less significant. The next two charts show that this efficiency gain is achieved with almost no loss in accuracy, as the difference in F1-scores between the original algorithm and our heuristic is less than 0.01.
    \item Fixed number of variables ($m = 50$) with increasing sample size. Figure~\ref{fig:simulated_results_fixm} shows that the original algorithm's search time grows linearly with the number of samples, while our heuristic's search time is again constant. The total runtime for our method is significantly lower across all sample sizes. The last two figures in the second row confirm that the accuracy is again comparable. It is worth noting that for very small datasets, the time saved by the faster search loop might not fully compensate for the initial overhead of the precomputation step. Our method demonstrates its primary advantages in the situations where the original algorithm becomes computationally intensive.
\end{itemize}

\subsection{Results on Real-World Datasets}
\label{sec:sp500}
\begin{table}[t]
\centering
\caption{Results for real datasets with ground truth in terms of F1-score and execution time (s).}
\begin{tabular}{l|c|c|c|c}
\hline
 & F1-score & \multicolumn{3}{c}{Execution Time (s)} \\
  &  & Pre-Cmp. & Ordering & Total\\
\hline
fMRI Before & 0.619 ± 0.139 & 0.000 & 0.342 & 0.524\\
fMRI After & 0.614 ± 0.133 & 0.009 & 0.010 & 0.221\\
\hline
Web1 Before & 0.262 & 0.000 & 0.374 & 2.588\\
Web1 After & 0.258 & 0.039 & 0.011 & 2.411\\
\hline
Web2 Before & 0.262 & 0.000 & 0.493 & 2.929\\
Web2 After & 0.286 & 0.041 & 0.012 & 2.351\\
\hline
Antivirus1 Before & 0.202 & 0.000 & 0.416 & 0.772\\
Antivirus1 After & 0.211 & 0.024 & 0.007 & 0.369\\
\hline
Antivirus2 Before & 0.205 & 0.000 & 0.364 & 0.734\\
Antivirus2 After & 0.205 & 0.026 & 0.001 & 0.333\\
\hline
\end{tabular}
\label{tab:web_comparison}
\end{table}

\begin{table}[t]
\centering
\caption{Execution time (seconds) and speedup on the S\&P500 dataset. `Original\textsubscript{1}' refers to the CPU version from \cite{ikeuchi2023python}, `GPU\textsubscript{2}' to the GPU version from \cite{akinwande2024acceleratedlingam}, and `Heuristic CPU\textsubscript{3}' is our version.}
\begin{tabular}{l r r r r r}
\hline
& \multicolumn{3}{c}{Execution Time (s)} & \multicolumn{2}{c}{Speed-Up} \\
\hline
Design & Original CPU\textsubscript{1} & GPU\textsubscript{2} & Proposed CPU\textsubscript{3} & S\textsubscript{13} & S\textsubscript{23} \\
\hline
$N_\textrm{variables} = 25$ & 4.03 & 8.83 & 1.88 & 2.14x & 4.69x \\
$N_\textrm{variables} = 50$ & 27.31 & 32.89 & 8.70 & 3.14x & 3.78x \\
$N_\textrm{variables} = 100$ & 230.06 & 168.04 & 44.54 & 5.17x & 3.77x \\
$N_\textrm{variables} = 200$ & 1660.57 & 1030.17 & 226.80 & 7.32x & 4.54x \\
$N_\textrm{variables} = 400$ & 21404.76 & 7291.09 & 1601.80 & 13.36x & 4.55x \\
\hline
\end{tabular}
\label{tab:sp500_comparison1}
\end{table}

To validate the effectiveness of our heuristic in practical scenarios, it is tested on several real-world benchmark datasets.
\begin{itemize}
\item Real-world data with ground-truth: We evaluate the heuristic on an fMRI dataset from neuroscience \cite{assaad2022survey} and IT monitoring datasets from \cite{ait-bachir2023case}. As shown in Table \ref{tab:web_comparison}, our method achieved a significant speedup on the fMRI dataset, reducing the total execution time by more than half. This efficiency gain came with a negligible change in the F1-score, which remained well within the standard deviation of the original method's performance. On the IT monitoring dataset, our method again consistently reduced execution times. The results on the Antivirus1 dataset show a significant reduction in runtime and a slight improvement in F1-score. However, we cannot rule out the possibility that the F1-score improvement is due to random variation or other factors not captured in our experiment. Further investigation is needed to confirm the significance of these findings.

\item Real-world datasets without ground-truth: To test scalability on a challenging, high-dimensional problem, we use S\&P500 stock data \cite{hyvarinen2010estimation}. We benchmark our CPU version against the standard CPU implementation from the \texttt{lingam} package \cite{ikeuchi2023python} and a GPU-accelerated version of the original algorithm \cite{akinwande2024acceleratedlingam}. The results are shown in Table \ref{tab:sp500_comparison1}. The performance advantage of our method grows dramatically with the number of variables. For 400 variables, our heuristic is 13.36 times faster than the original CPU algorithm and 4.55 times faster than the GPU implementation. The original algorithm took nearly 6 hours to run, while our heuristic finished in under 27 minutes on the same hardware. This shows that for achieving scalability, a proper algorithmic design can be more effective than hardware acceleration of an inefficient algorithm. The accessibility of running such large-scale analyses on a standard laptop is a key practical outcome of our work.
\end{itemize}

\section{Discussion}

Our work successfully demonstrates the value of a heuristic approach to a computationally difficult problem. This section contextualizes our contribution, discusses the inherent limitations of our method, and analyzes the practical trade-offs related to scalability and system resources.

\subsection{Primary Contribution in Context}
The main contribution of this work is the design and validation of a new point in the design space for causal discovery algorithms. While precomputation is a known optimization technique, its application to DirectLiNGAM required a deliberate algorithmic modification: the omission of the iterative data refinement step. The novelty of our contribution is not the act of precomputation itself, but the empirical demonstration that this specific and aggressive approximation is highly effective within the VarLiNGAM context.

Our findings position this algorithmic approximation as a practical alternative to purely hardware-centric acceleration. Our efficient algorithm on standard hardware can outperform the original algorithm running on a specialized processor like a GPU. For example, in the experiments described in Section~\ref{sec:sp500} with the S\&P500 stock data, the proposed approach running on the CPU achieves up to 4.55 times speed-up over the original VarLiNGAM running on the GPU. This suggests that for practitioners without access to high-performance computing resources, exploring algorithmic heuristics can be a more accessible and effective path to achieving scalability. The impact of this work is most significant for users with standard, commodity hardware, as it enables them to perform large-scale causal discovery that is previously infeasible.

\subsection{Limitations of the Heuristic Approach}

The primary limitation is the heuristic nature of the algorithm. By omitting the iterative residualization step, we lose the theoretical guarantee of correctness that the original DirectLiNGAM algorithm provides. Our experiments suggest that the accuracy loss is minimal in many practical cases. However, there may exist specific data generating processes, perhaps with very subtle causal links that are obscured by stronger, indirect effects, where our heuristic could fail to find the correct causal ordering while the original algorithm would succeed. Characterizing the theoretical bounds of when this approximation holds is a non-trivial problem and an important direction for future research.

Moreover, this paper focuses on the algorithmic trade-off between speed and accuracy. It does not attempt to explain the domain-specific mechanisms behind the causal relationships discovered in the real-world datasets. The tool we develop is intended to be used by domain experts who can provide the necessary context and interpretation for the resulting causal graphs.

\subsection{Scalability and Resource Trade-offs}
Our method achieves its speedup by trading computational time for memory. The \(O(m^2)\) space complexity for storing the precomputed residual entropies is a key aspect of this trade-off. For the datasets used in our experiments (up to 400 variables), this memory footprint is minor on modern systems. However, for systems with severely limited memory resources, or for problems with an extremely large number of variables (many thousands), this could become a bottleneck. In such scenarios, memory optimization techniques could be considered, such as using lower-precision floating-point numbers or developing a hybrid strategy that only precomputes a subset of the most frequently accessed values.

Regarding performance in high-dimensional settings, our heuristic offers a significant improvement. However, its scalability is not infinite. While we have reduced the dependency on the number of samples \(n\), the complexity still includes an \(O(m^3)\) term from the search phase. As the number of variables \(m\) grows into the thousands, this term will eventually become the new computational bottleneck, particularly for datasets where \(m >> n\). Even so, improving the complexity from \(O(m^3 n)\) to \(O(m^2 n + m^3)\) represents a substantial step forward in making higher-dimensional analysis more tractable.

\section{Conclusion}

This study introduced and validated a novel heuristic approximation of the VarLiNGAM algorithm, designed to overcome the computational barriers that limit the use of causal discovery on large-scale time-series data. By making a deliberate algorithmic change to the core DirectLiNGAM estimator, specifically by omitting the iterative data refinement step, we enable an efficient precomputation strategy. This modification reduces the computational complexity significantly, resulting in major speedups that make analysis of datasets with hundreds of variables feasible on a standard laptop.

Our evaluation demonstrated that this gain in efficiency comes at a negligible cost to empirical accuracy across a variety of synthetic and real-world problems. This work highlights that for certain classes of complex algorithms, a well-designed algorithmic approximation can be a more effective and accessible path to scalability than pure hardware acceleration. Future work could include exploring the theoretical conditions under which our heuristic is guaranteed to match the output of the original algorithm. Additionally, similar precomputation strategies could be investigated for other parts of the VarLiNGAM workflow, such as the pruning stage, or one could explore whether patterns in the precomputed entropy matrices could be used to guide the causal search even more efficiently.

\bibliographystyle{plain}
\bibliography{reference}

\end{document}